%% file: main.tex
\documentclass[journal]{vgtc}                     

\onlineid{0}

\vgtccategory{Research}

\vgtcpapertype{Analytics \& Decisions}

\title{Token Signatures of Code: Comparing Coding Behaviors Across Large Language Models}

\author{%
  Junpeng Wang, Yuzhong Chen, Menghai Pan, Uday Singh Saini, and Yiwei Cai
}

\authorfooter{
  \item
  	J. Wang, Y. Chen, M. Pan, U. Saini, Y. Cai are with Visa Research.
  	E-mail: \{junpenwa, yuzchen, menpan, udasaini, yicai\}@visa.com

}

\input{tex/0abstract}

\teaser{
  \centering
  \includegraphics[width=\linewidth, alt={The CLIC visual analytics system, composed of three coordinated views: the Metric View (matrix and scatterplot), the Decision Tree View (node-link diagram with feature importance), and the Feature Ablation View (bar and area charts showing progressive feature removal).}]{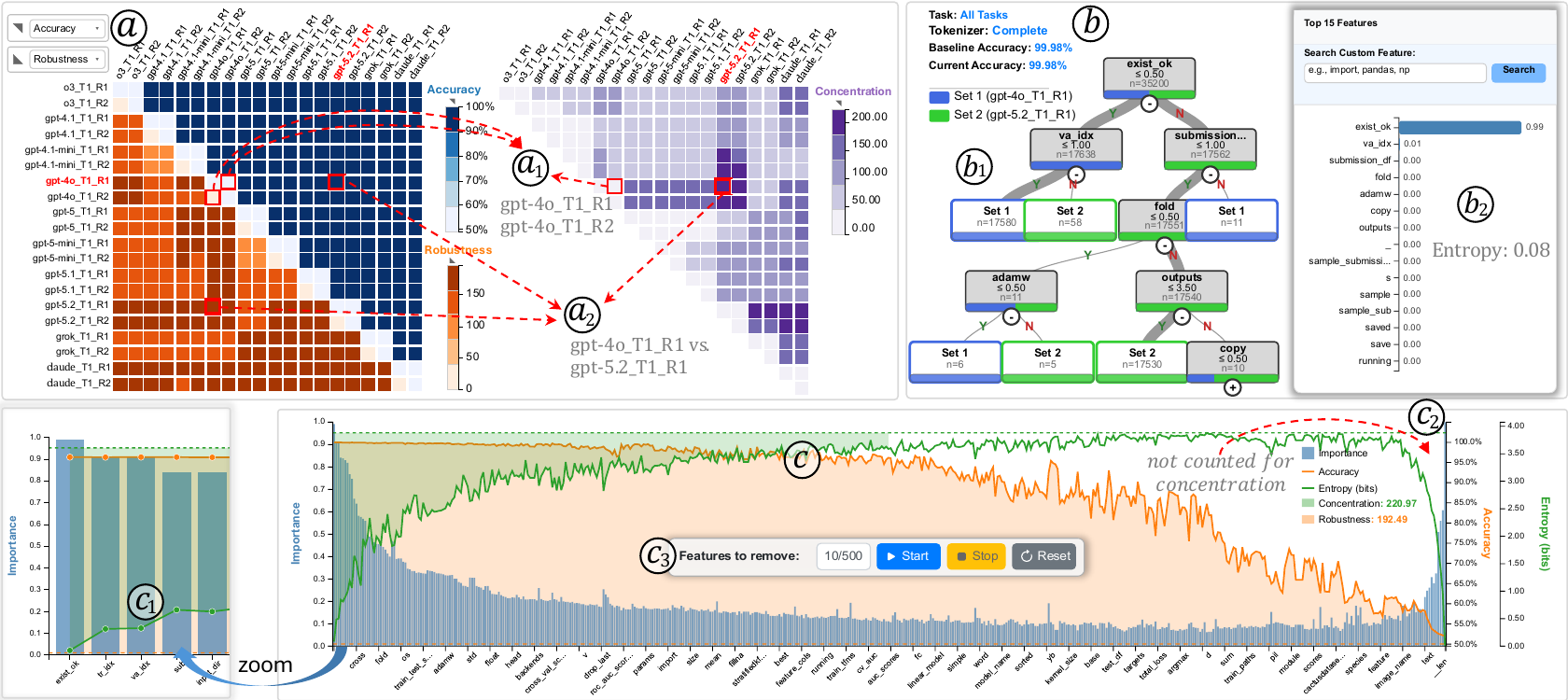}
  \vspace{-0.23in}
  \caption{\sysname{} is composed of three coordinated views. The \metricview{} (a) alternates between a matrix and scatterplot visualization to present three metrics quantifying the separability of LLM pairs. The \treeview{} (b) visualizes the trained decision tree used to separate two code sets, along with the top discriminative features. The \ablationview{} (c) records the dynamics when progressively removing the most important features one after another and supports time travel back to an iteration of interest.
  }
  \label{fig:system}
}

\newcommand{\sysname}{\textit{CLIC}}

\graphicspath{{figs/}{figures/}{pictures/}{images/}{./}} 

\usepackage{xcolor}
\usepackage{tabu}                      
\usepackage{booktabs}                  
\usepackage{multirow}
\usepackage{lipsum}                    
\usepackage{mwe}                       
\usepackage{enumitem}
\usepackage{mathptmx}                  
\usepackage[most]{tcolorbox}
\usepackage{algorithm}
\usepackage{algpseudocode} 
\usepackage{amsmath}       
\usepackage{float}         
\usepackage{listings}
\algrenewcommand\algorithmiccomment[1]{\textcolor{ForestGreen}{\(\triangleright\) #1}}
\usepackage{amssymb}  
\usepackage{tikz}
\newcommand{\llblacktriangle}{\mathord{\text{%
  \tikz[baseline=\dp\strutbox]{\fill[gray] (0pt,0pt)--({\ht\strutbox+\dp\strutbox},0pt)--(0pt,{\ht\strutbox+\dp\strutbox})--cycle;}}}}
\newcommand{\urblacktriangle}{\mathord{\text{%
  \tikz[baseline=\dp\strutbox]{\fill[gray] (0pt,{\ht\strutbox+\dp\strutbox})--({\ht\strutbox+\dp\strutbox},{\ht\strutbox+\dp\strutbox})--({\ht\strutbox+\dp\strutbox},0pt)--cycle;}}}}

\newcommand{\treeview}{\textit{Decision Tree View}}
\newcommand{\metricview}{\textit{Metric View}}
\newcommand{\ablationview}{\textit{Feature Ablation View}}
\newcommand{\codeview}{\textit{Code Context View}}

\definecolor{ronecolor}{rgb}{0.937, 0.745, 0.173}
\newtcbox{\reqboxone}{on line,
  colframe=ronecolor,    
  colback=ronecolor,
  coltext=white,
  boxrule=0.4pt,
  arc=3pt,
  boxsep=0.1pt,
  left=2pt,right=2pt,top=1pt,bottom=1pt,
}

\definecolor{codecolor}{rgb}{0.518, 0.671, 0.314}
\newtcbox{\reqboxtwo}{on line,
  colframe=codecolor,    
  colback=codecolor,
  coltext=white,
  boxrule=0.4pt,
  arc=3pt,
  boxsep=0.1pt,
  left=2pt,right=2pt,top=1pt,bottom=1pt,
}

\definecolor{analyzecolor}{rgb}{0.8549, 0.4667, 0.2588}
\newtcbox{\reqboxthree}{on line,
  colframe=analyzecolor, 
  colback=analyzecolor,
  coltext=white,
  boxrule=0.4pt,
  arc=3pt,
  boxsep=0.1pt,
  left=2pt,right=2pt,top=1pt,bottom=1pt,
}

\definecolor{sharpcolor}{rgb}{0.871, 0.945, 0.827}
\newtcbox{\sharpcascading}{on line,
  colframe=sharpcolor, 
  colback=sharpcolor,
  coltext=black,
  boxrule=0.4pt,
  arc=.5pt,
  boxsep=0.1pt,
  left=.1pt,right=.1pt,top=1pt,bottom=1pt,
}

\definecolor{deepcolor}{rgb}{0.9686, 0.8941, 0.847}
\newtcbox{\deepdistributed}{on line,
  colframe=deepcolor, 
  colback=deepcolor,
  coltext=black,
  boxrule=0.4pt,
  arc=.5pt,
  boxsep=0.1pt,
  left=.1pt,right=.1pt,top=1pt,bottom=1pt,
}

\definecolor{fragilecolor}{rgb}{0.8706, 0.8706, 0.9882}
\newtcbox{\fragilediffuse}{on line,
  colframe=fragilecolor, 
  colback=fragilecolor,
  coltext=black,
  boxrule=0.4pt,
  arc=.5pt,
  boxsep=0.1pt,
  left=.1pt,right=.1pt,top=1pt,bottom=1pt,
}

\definecolor{lexicalcolor}{rgb}{0.9247, 0.9247, 0.9247}
\newtcbox{\lexicalbrittle}{on line,
  colframe=lexicalcolor, 
  colback=lexicalcolor,
  coltext=black,
  boxrule=0.4pt,
  arc=.5pt,
  boxsep=0.1pt,
  left=.1pt,right=.1pt,top=1pt,bottom=1pt,
}

\definecolor{sumcolor}{rgb}{0.2824, 0.6118, 0.8157}
\newtcbox{\sumllmbox}{on line,
  colframe=black,        
  colback=codecolor,
  coltext=white,        
  boxrule=0.2pt,        
  arc=1.5pt,              
  boxsep=0.1pt,
  left=2pt,right=2pt,top=1pt,bottom=1pt,
}

\definecolor{tomato}{rgb}{1.0, 0.39, 0.28}
\definecolor{limegreen}{rgb}{0.196, 0.804, 0.196}
\definecolor{lightlimegreen}{rgb}{0.68, 1.0, 0.68}  
\definecolor{lighttomato}{rgb}{1.0, 0.7, 0.6}      
\newtcbox{\simcircle}{on line,
  colframe=black,        
  colback=lightlimegreen,     
  coltext=black,         
  boxrule=0.2pt,         
  arc=4pt,               
  boxsep=0.1pt,
  left=1.85pt,right=1.85pt,top=1pt,bottom=1pt,
}

\newtcbox{\diffcircle}{on line,
  colframe=black,        
  colback=lighttomato,     
  coltext=black,         
  boxrule=0.2pt,         
  arc=4pt,               
  boxsep=0.1pt,
  left=1.85pt,right=1.85pt,top=1pt,bottom=1pt,
}

\begin{document}


\firstsection{Introduction}

\maketitle

\input{tex/1introduction}
\input{tex/2relatedwork}

\input{tex/3requirements}

\input{tex/4generation}

\input{tex/5comparison}

\input{tex/6system}

\input{tex/7casestudy}

\input{tex/8feedback}

\input{tex/9discussion}
\input{tex/10conclusion}




\bibliographystyle{abbrv-doi-hyperref}

\bibliography{template}








\end{document}

%% file: tex/0abstract.tex
\abstract{%
 The evaluation of large language models (LLMs) on coding tasks has primarily focused on performance metrics such as pass@$k$. As LLMs continue to advance, many models now meet baseline performance requirements, reducing the discriminative power of performance-based evaluation alone. Yet a key question remains largely unexplored: \textit{how do LLMs differ in their \textbf{coding behavior}?} We propose \sysname{} (Code Learning for Identification and Comparison), a visual analytics approach that characterizes LLM coding behavior through token-frequency analysis. \sysname{} represents each code sample as a feature vector of token frequencies and trains an interpretable decision tree to separate two LLMs' code sets. Beyond classification accuracy, we define two new metrics: \textit{robustness}, which measures whether the two LLMs remain distinguishable as their most-discriminative tokens are progressively removed, and \textit{concentration}, which measures whether the difference is driven by a few dominant tokens or spread across many. Interpreting numerous pairwise comparisons---across LLM pairs, tasks, and tokenization levels---and tracing the full analytical chain form an inherently multi-scale, hypothesis-driven exploration task. We therefore develop an interactive visual analytics system to navigate the comparison landscape, identify pairs of interest, and drill down into discriminative tokens and their code contexts. Case studies comparing 10 LLMs across 22 Kaggle ML tasks reveal actionable insights for LLM selection and prompt engineering.
}

\keywords{LLM Comparison, Agentic Coding, Artificial Intelligence, Visualization, Visual Analytics.}

%% file: tex/1introduction.tex
\makeatletter\protected@edef\@currentlabel{\thesection}\makeatother\label{sec:introduction}
With the rapid proliferation of LLMs, systematic comparison among them has become increasingly critical~\cite{kahng2024llm, kahng2024llm1, qin2024infobench, zeng2023evaluating, wang2026understanding}. Domain practitioners regularly ask: Which LLM adheres more closely to a prompt? Which is more cost-efficient in real-world use? To address these needs, numerous benchmarks~\cite{qin2024infobench, li2024salad, ivanov2024ai} have been proposed to evaluate LLMs across many dimensions. When it comes to LLM-powered coding tasks, however, most existing work remains narrowly focused on functional correctness or execution performance~\cite{chan2024mle, wang2025illuminating, jiang2024survey}. For example, MLE-Bench~\cite{chan2024mle} ranks LLMs and agentic coding frameworks by a performance metric, pass@$k$~\cite{chen2021evaluating}, on its leaderboard~\cite{mlebenchleaderboard}.

As LLM capabilities continue to advance, many models now meet baseline performance requirements on standard coding tasks, further reducing the discriminative power of \textit{\textbf{performance-based}} evaluations. Yet an equally important question remains largely unexplored: \textit{how do LLMs differ in their coding behavior?} For instance, among LLMs with similar performance, does one consistently produce more verbose code than another? Does one exhibit greater caution in file operations? Such \textit{\textbf{behavior-based}} differences, while often overlooked, can meaningfully affect code reliability, maintainability, and the quality of developer interaction. Understanding them is essential for accurately characterizing LLMs, guiding prompt engineering, and selecting LLMs. To make this concrete, consider two scenarios. \textit{First}, when iterating on an agentic coding framework---e.g., refining a prompt or adding a memory module~\cite{zhao2026memory}---developers need to know how the resulting code \emph{differs} from before, since the modification rarely manifests as a single scalar improvement and may shift the model's behavior in unintended ways. \textit{Second}, when multiple LLMs achieve comparable performance on a target task, the choice between them hinges on which model's coding behavior aligns best with the team's conventions, downstream tooling, and reliability needs. Both scenarios demand a principled, interpretable comparison of \emph{how} two LLMs code, not merely \emph{whether} they pass tests. They are also the daily concerns of our target users: \textit{\textbf{ML scientists and engineers} who use LLMs for code generation}.

To fill this research gap, we propose \underline{\textit{C}}ode \underline{\textit{L}}earning for \underline{\textit{I}}dentification and \underline{\textit{C}}omparison (\sysname{}). \sysname{} compares LLMs' coding behavior by analyzing token frequencies in LLM-generated code. Concretely, it represents each code sample as a feature vector of token frequencies (computed at five distinct semantic levels). Two collections of code from two LLMs thus become two sets of feature vectors. We train a binary decision tree to separate them, and its \textit{accuracy} quantifies how distinguishable the two LLMs' coding behaviors are. The cornerstone of this pipeline is the token-frequency representation itself. While \textit{\textbf{token frequency}} may at first appear to be a low-level lexical property, in our setting, it is in fact a behaviorally rich, well-validated representation that surfaces concrete coding behaviors. We focus on it for four reasons. \textit{First}, tokens are the atomic units produced by LLMs, and their frequencies directly reflect a model's generation preferences. \textit{Second}, many code tokens carry clear semantics, and their frequency differences correspond to distinct coding behaviors, e.g., a high occurrence of the \texttt{print} token reflects \emph{verbose} code. \textit{Third}, token-frequency analysis is the canonical feature representation of NLP authorship attribution and code stylometry~\cite{caliskan2015deanon, bisztray2025know}, repeatedly shown to reliably distinguish source-code authors. This long-validated stylometric tradition grounds our use of token frequency as a meaningful lens on coding behavior. \textit{Fourth}, token-frequency analysis is intuitive and easy to communicate, enabling clear and interpretable comparisons across LLMs.

The decision tree \textit{\textbf{accuracy}} alone, however, does not fully characterize coding behavior differences. Two LLM pairs with the same \textit{accuracy} can differ substantially in how that separability is structured. One pair's separability may reside entirely in a single token: removing it causes the \textit{accuracy} to collapse to ${\sim}50\%$ (random guessing). The other pair's separability may persist even after the top token is removed, as a new token rises to take its place. This contrast motivates our \textit{\textbf{robustness}} metric, which measures whether two LLMs remain distinguishable as their most-discriminative tokens are progressively removed. A separate question concerns how the discriminative signal is \emph{distributed} at each step: is it driven by a single dominant token, or spread across many? This motivates our \textit{\textbf{concentration}} metric, which measures the skewness of the token-importance distribution---high when a few tokens dominate, low when many tokens contribute roughly equally.

While the three metrics quantify behavioral differences, applying them at scale---across many LLM pairs, coding tasks, and tokenization levels---yields a multi-granular, hypothesis-driven workflow: an analyst first explores the comparison landscape to identify pairs of interest, then drills into the discriminative tokens, and finally examines code samples to validate a hypothesis. Findings at one level raise new questions at the next, demanding coordinated, interactive views rather than any single static representation~\cite{cook2005illuminating,roberts2007state}. We therefore develop a visual analytics system (Fig.~\ref{fig:system}) to support this exploration. The system presents metrics across LLM pairs through matrix and scatterplot visualizations (\metricview{}), exposes the trained decision tree via a node-link diagram (\treeview{}), and illustrates feature-importance dynamics through coordinated bar and area charts (\ablationview{}).

We conduct multiple case studies comparing 10 LLMs on Python code generated for 22 Kaggle ML competitions, showing how \sysname{} reveals behavioral differences that inform LLM selection and steer prompt engineering. 
Although we focus on Python code and ML tasks, the proposed methodology is task-agnostic and can be directly extended to other programming languages, a direction we leave for future work.

To summarize, our contributions are threefold:
\begin{enumerate}[topsep=0pt, partopsep=0pt, labelsep=0.1cm, itemsep=-0.07cm]
    \item We propose five levels of code tokenization and a learning-to-compare algorithm to compare LLMs' code at each level.
    \item We introduce two new metrics, \textit{robustness} and \textit{concentration}, to thoroughly characterize the separability between two LLMs.
    \item We develop a visual analytics system to explore the LLM comparison results and steer the drill-down exploration process.
\end{enumerate}

%% file: tex/2relatedwork.tex
\section{Related Work}
\noindent \textbf{LLM Code Generation and Benchmark.}
LLMs have shown remarkable capability on coding tasks~\cite{dong2025survey}, giving rise to a rich ecosystem of coding frameworks~\cite{aide, nam2025mle,yang2025rdagent, toledo2025ai, liu2025ml} and benchmarks~\cite{chen2021evaluating,jimenez2024swebench,chan2024mle,wijk2024re} to evaluate them. Early work introduced dedicated code LLMs such as Codex~\cite{chen2021evaluating} and CodeT5~\cite{wang2021codet5}, and tools like GitHub Copilot~\cite{githubcopilot} brought LLM-powered code generation to everyday practice. Building on these foundations, more recent agentic frameworks automate multi-step coding workflows. For example, OpenHands~\cite{wang2024openhands} introduces a chain-based iterative solution search. AIDE~\cite{aide} extends it to a tree-based search, which enables the agent to alternately explore multiple solution branches. R\&D-Agent~\cite{yang2025rdagent} further introduces a graph-based solution search, enabling solution fusion to leverage knowledge gained across branches.
For evaluation, HumanEval~\cite{chen2021evaluating} introduces 164 hand-written programming problems and the pass@$k$ metric, establishing the dominant evaluation paradigm of functional correctness. SWE-bench~\cite{jimenez2024swebench} extends this to real-world software engineering by testing LLMs on resolving GitHub issues, while MLE-Bench~\cite{chan2024mle}, which we use in this work, evaluates agentic LLM coding on Kaggle ML competitions. A comprehensive survey by Jiang et al.~\cite{jiang2024survey} catalogs the broader landscape of evaluation across code generation, completion, and repair. 
Despite this breadth, these benchmarks~\cite{austin2021program, chan2024mle, chen2021evaluating,jimenez2024swebench} and frameworks~\cite{aide,yang2025rdagent,fang2025mlzero} share a common focus: whether generated code executes correctly. \textit{How LLMs differ in the way they write code---their stylistic and behavioral tendencies---remains largely unexamined.}

\noindent \textbf{LLM Visual Comparison.} The growing adoption of LLMs has motivated a body of work on visually comparing their behaviors~\cite{chen2024viseval,sevastjanova2023visual, pan2025vis,wang2024visualization,coscia2024iscore}. 
LLM Comparator~\cite{kahng2024llm, kahng2024llm1} juxtaposes LLM responses to help analysts discover cases where one model performs better than another.
VisEval~\cite{chen2024viseval} evaluates and compares multiple LLMs on the task of translating natural-language queries over tabular data into executable visualization specifications, revealing common failure patterns such as incorrect data mapping and poor design choices.
EvalLM~\cite{kim2024evallm} lets users define custom criteria and leverage LLMs to evaluate model outputs, supporting iterative prompt refinement. ChainForge~\cite{arawjo2024chainforge} enables cross-LLM prompt comparison through an interactive interface, organizing responses in structured views to support hypothesis testing. 
\textit{While these systems focus on comparing LLM capabilities in language or image generation, none focuses on comparing their coding capabilities.} 
Most recently, Wang et al.~\cite{wang2025illuminating} developed a visual analytics system to study and improve LLM-powered coding agents, exposing agent behavior through tree-structured exploration and code examination. Their work focuses on distinguishing LLMs in their iterative coding strategy exploration process. \textit{Our work instead focuses on coding style differences across LLMs, particularly their token usage patterns.}

\noindent \textbf{VIS4AI and Comparative Analysis.} Leveraging visualization to make AI models more interpretable (VIS4AI~\cite{wang2024visual,liu2025visualization,hohman2018visual}) has been a prevalent research direction. Many works have demonstrated that visualization can effectively help humans better understand~\cite{wang2018ganviz, liu2016towards,li2023does}, diagnose~\cite{rathore2024verb,wang2019deepvid,wexler2019if, li2020cnnpruner, strobelt2018s}, improve~\cite{gou2020vatld,bilal2017convolutional, wang2018dqnviz}, and steer~\cite{yang2020interactive,ming2019protosteer,li2024visual} AI models. Among these works, two groups are closely related to ours. The first is interpretable tree visualization. Decision trees are inherently interpretable and there are many visualizations designed for them~\cite{muhlbacher2017treepod,le2021treeheatr}. For example, BaobabView~\cite{van2011baobabview} is a visual analytics system for interactively constructing, editing, and analyzing decision trees. It uses a flow-based tree visualization where link width and color encode how data instances of different classes move through the splits.
Later works adapt node-link diagrams to visualize decision trees and facilitate the analysis of an ensemble of trees~\cite{liu2017visual, wang2021investigating,sondag2025cluster}. \textit{We employ a similar visualization to present the decision tree for its intuitive clarity.} The second related VIS4AI direction is visual comparative analysis~\cite{gleicher2011visual,gleicher2017considerations}. The most closely related to our work is learning-from-disagreement (LFD)~\cite{wang2022learning,wang2022learning1}, which trains a classifier to distinguish instances on which two ML models disagree, and probes important features to characterize the models' relative strengths and weaknesses. \textit{Similar to LFD, our work also uses a binary classifier to distinguish two LLMs' coding behavior, but through the lens of their token usage. Furthermore, we introduce two new metrics, \textbf{robustness} and \textbf{concentration}, built on the trained classifier to more comprehensively quantify the separability.}

%% file: tex/3requirements.tex
\section{Requirement Analysis}
\label{sec:requirement}
\sysname{} was developed through a user-centered design process in close collaboration with three ML scientists (target users). All hold Ph.D. degrees in computer science and have 3--5 years of full-time industry experience. Their primary responsibilities involve building and deploying ML models in the financial industry. In their work, they extensively use LLMs for code generation and are well-versed in state-of-the-art (SOTA) agentic coding frameworks. Over two months, we conducted weekly meetings to iteratively elicit their pain points in LLM selection and generated-code comparison. Alongside this user-centered process, our requirement analysis was also actively informed by three threads of prior literature, each directly shaping one core aspect of the final requirements. \textit{First}, classifier-based two-sample testing~\cite{lopez2017c2st} and recent LLM-code stylometry~\cite{bisztray2025know} motivated us to formulate requirements for measuring code separability with \emph{quantifiable metrics}. \textit{Second}, corpus-contrastive analyses~\cite{monroe2008fightin, kessler2017scattertext} drove us to formulate requirements for attributing the observed separability back to \emph{individual code tokens} and their usage contexts. \textit{Third}, code-stylometry research~\cite{caliskan2015deanon} led us to formulate requirements for decomposing code into distinct \emph{semantic levels} and analyzing each level separately. Synthesizing the experts' pain points with these literature insights and broader prior work on LLM code generation~\cite{dong2025survey, twist2025llms, wang2025illuminating}, we identified three key requirements.

\begin{itemize}[label={},topsep=0pt, partopsep=0pt, leftmargin=0cm, labelsep=0.1cm, itemsep=-0.06cm]
\item \textbf{\reqboxone{R1} Quantifying Code Separability}: To characterize how LLMs differ in their generated code, the experts preferred quantitative metrics to holistically measure the \emph{separability} between two code sets, i.e., how easily the originating LLM can be inferred from the code and whether a few features dominate the separability. Moreover, since practitioners can easily modify LLM prompts to regenerate code, it is necessary to assess the robustness of this separability. Thus, we need:
\begin{itemize}[label=$\diamond$, topsep=-15pt, partopsep=-15pt, leftmargin=0.3cm, labelsep=0.1cm, itemsep=-0.02cm]
    \item \reqboxone{R1.1} An intuitive metric quantifying the \textit{separability} between two code sets, reflecting how distinguishable the underlying LLMs are.
    \item \reqboxone{R1.2} A metric that captures the \textit{robustness} of the observed separability under progressive ablation of the key discriminative features.
    \item \reqboxone{R1.3} A metric that characterizes the \textit{skewness} of the discriminative feature distribution, i.e., how the separability is structured internally.
\end{itemize}

\item \textbf{\reqboxtwo{R2} Identifying Key Discriminative Features}: Beyond global separability measures, it is necessary to pinpoint the specific features (code tokens) that drive the observed differences. Understanding which tokens are more discriminative and how they collectively contribute to the separability enables more targeted LLM characterization and informed prompt engineering. This requires \sysname{} to:
\begin{itemize}[label=$\diamond$,topsep=-15pt, partopsep=-15pt, leftmargin=0.3cm, labelsep=0.1cm, itemsep=-0.02cm]
    \item \reqboxtwo{R2.1} Identify the features that contribute the most to the separation of two code sets and the code contexts where the features were used.
    \item \reqboxtwo{R2.2} Reveal the distributional pattern of discriminative features, characterizing whether the separability is driven by a dominant feature or emerges from the joint effect of multiple features.
    \item \reqboxtwo{R2.3} Support what-if analysis~\cite{wexler2019if} by excluding the most important features, uncovering features of equivalent discriminative power masked by more dominant ones.
\end{itemize}

\item \textbf{\reqboxthree{R3} Multi-Level Code Analysis}: A piece of code is composed of distinct semantic levels, e.g., comments, functional code, and imported packages. 
Each reflects a unique aspect of LLM coding behavior, and the separability may disappear when one level is peeled off. It is therefore necessary to attribute observed differences to specific levels for fine-grained characterization. For instance, divergences may primarily reside in code comments, reflecting differences in natural language capability. One LLM may consistently favor tree-based approaches while another defaults to neural network solutions, reflecting differences in model preference. To reveal this, \sysname{} needs to:
\begin{itemize}[label=$\diamond$,topsep=-15pt, partopsep=-15pt, leftmargin=0.3cm, labelsep=0.1cm, itemsep=-0.02cm]
    \item \reqboxthree{R3.1} Decompose code into distinct semantic levels and quantify the separability and key discriminative features at each level.
    \item \reqboxthree{R3.2} Compare separability across levels to identify which aspects of code generation most distinguish the two LLMs.
\end{itemize}
\end{itemize}

%% file: tex/4generation.tex
\section{Code Generation}
\label{sec:codegeneration}

\textbf{Coding Tasks.}
We investigate the differences in LLMs' coding behaviors by comparing their generated code. To ground the comparison in a concrete and reproducible setting, we use Python code generation for ML tasks as our study domain. Rather than relying on a single task, we evaluate on 22 ML tasks (22 Kaggle competitions from MLE-Bench~\cite{chan2024mle}) spanning distinct data modalities (tabular, image, text, and audio) and a range of difficulty levels. This internal diversity exposes \sysname{} to qualitatively different problems within a single, well-validated benchmark~\cite{chan2024mle}, and the per-task analysis (in Appendix) shows that the discriminative tokens shift task-by-task rather than collapsing into a single task-specific signal. We emphasize that this is a methodological choice: \sysname{}'s comparison mechanism makes no assumptions specific to ML tasks, and it applies directly to other software-development scenarios, which we discuss as future work in Sec.~\ref{sec:discussion}.

\textbf{Coding Agents.}
To isolate differences attributable to the LLM rather than the agentic coding frameworks, we use a single coding agent throughout this work, varying only the backend LLM.
Specifically, we choose AI-Driven Exploration (AIDE)~\cite{aide, aidecode}, as it is open-source and achieves SOTA performance.
It takes natural language descriptions of an ML problem as input and generates code to iteratively solve it. Within each iteration, the agent can \textit{draft} an initial solution, \textit{debug} existing buggy code, or \textit{improve} functional code. Since the latter two depend on code from prior iterations and thus introduce variability beyond the LLM itself, we only use the \textit{draft} operation to ensure all LLMs generate code under identical conditions.

\textbf{LLM Settings.}
Our work uses 10 LLMs: \texttt{o3}, \texttt{gpt-4o}, \texttt{gpt-4.1}, \texttt{gpt-4.1-mini}, \texttt{gpt-5}, \texttt{gpt-5-mini}, \texttt{gpt-5.1}, \texttt{gpt-5.2}, \texttt{grok-4} (\texttt{grok} for short), and \texttt{claude-sonnet-4.5} (\texttt{claude} for short)~\footnote{All trademarks are the property of their respective owners, are used for identification purposes only, and do not necessarily imply product endorsement or affiliation with Visa.}. They are chosen along two complementary axes that maximize the comparative leverage of \sysname{}---(i)~\emph{diversity across providers} (OpenAI, xAI, and Anthropic), exposing stylistic differences attributable to different model families and training regimes, and (ii)~\emph{similarity within providers} (e.g., \texttt{gpt-5} vs.\ \texttt{gpt-5-mini}), exposing fine-grained behavioral differences between closely related models. Given the rapid pace at which frontier LLMs are released, exhaustive coverage in any single study is intrinsically infeasible. The contribution of \sysname{} is methodological rather than a benchmark of LLM coverage: the comparison mechanism is agnostic to the specific LLMs being compared and can be readily transferred to any pair of code corpora.
For each LLM, AIDE attempts to solve the 22 Kaggle competitions $1,000$ times from scratch with \textit{\textbf{identical}} prompts, generating $1,000$ code samples per task. We call this a single LLM \textbf{run}. We perform two runs per LLM to verify behavioral consistency: if an LLM's coding style is stable, the two runs should be hard to distinguish from each other. By default, we set the temperature to 1 for all LLMs. In total, our experiment generates $10\ (LLMs) {\times} 22\ (tasks) {\times} 2\ (runs) {\times} 1,000\ (code) {=} 440{,}000$ code samples, which form the corpus analyzed later in Sec.~\ref{sec:all-tasks}.

%% file: tex/5comparison.tex
\section{Code Comparison with a Decision Tree}
Our approach represents each code sample as a token-frequency vector and trains an interpretable binary classifier to separate two LLMs' code sets, yielding both \emph{a separability score} and \emph{the most discriminative tokens}. Concretely, given $2{\times}1,000$ code samples from two LLM runs, we tokenize each sample\footnote{Breaking down the code into a set of atomic words. The terms \textit{\textbf{word}}, \textit{\textbf{token}}, and \textit{\textbf{feature}} are used interchangeably. Note that our tokenizer differs from LLM tokenizers, which may split a single word into multiple sub-word tokens.} and compute the frequencies of the top-500 tokens (empirically determined), representing each sample as a 500-dimensional feature vector. This high-dimensional representation can be analyzed by many paradigms, e.g., dimensionality reduction, embedding-based comparison, or statistical feature analysis. We analyze it with an interpretable discriminative classifier, because it is superior to others in directly returning the discriminative tokens that drive the separation between two code sets (\reqboxtwo{R2}). Within this paradigm, we choose to use decision tree models for three deliberate reasons:
\begin{enumerate}[topsep=0pt, partopsep=0pt, itemsep=-0.09cm]
\item \textit{Faithful, glass-box interpretability}: unlike post-hoc surrogates, a decision tree is itself the model---every prediction can be traced exactly to a sequence of token-frequency conditions, which is essential for attributing differences to specific lexical evidence.
\item \textit{Established precedent for analogous tasks}: decision trees have been repeatedly adopted as the interpretable backbone for visual analytics systems~\cite{wang2022learning,ming2018rulematrix,yuan2022hierarchical}, validating their suitability for the contrastive, token-level interpretation our work requires.
\item \textit{Alignment with target users}: our experts (consulted during the requirement analysis stage) explicitly indicated familiarity and trust in tree-based reasoning based on their experiments, reducing the cognitive overhead of adopting \sysname{} in real workflows.
\end{enumerate}
We emphasize, however, that this choice is intentionally \emph{modular}: the binary classifier requires only a model whose decisions can be locally attributed to input features, so any inherently interpretable classifier (e.g., linear models or generalized additive models) can be substituted. We leave a full comparison across these alternatives as future work.

We train a decision tree on the $2{,}000$ vectors using an 80/20 stratified train/test split (see pseudocode in the Appendix). All reported \textit{accuracy} values in this paper are evaluated on the held-out test set, ensuring the reported separability reflects generalization without overfitting.

The trained decision tree yields two key outputs: its \textit{accuracy} quantifies how distinguishable the two LLMs' coding behaviors are, and its \textit{token importance scores} (the tokens' cumulative Gini impurity decrease across all tree nodes) reveal which tokens drive the separation.

\subsection{Three Metrics Quantifying Separability}
\label{sec:metric}
We introduce three complementary metrics---\textit{accuracy}, \textit{robustness}, and \textit{concentration}---each capturing a different dimension of separability.

\textbf{Accuracy.}
The classifier's test accuracy intuitively reflects the separability. \textit{\textbf{A higher accuracy}} indicates that the two code sets are more distinct and easily separable, while \textit{\textbf{an accuracy near 50\%}} suggests the code sets are similar and difficult to distinguish (\reqboxone{R1.1}). We choose accuracy because it aligns naturally with our balanced classifier setting, in which the two compared code sets contribute a comparable number of instances and play \emph{symmetric} roles---neither is treated as a privileged ``positive'' class. Metrics such as precision, recall, and F1-score additionally encode \emph{which} class is mistaken for which, but this asymmetric information is not meaningful for our goal of quantifying separability between two interchangeable code sets. Accuracy is therefore selected not because it is universally superior, but because it provides the most parsimonious and interpretable aggregate. This choice is also \emph{modular} within \sysname{} and can be substituted by other metrics when necessary.

\textbf{Robustness.}
The accuracy metric surfaces only the initial separability, but not its depth. Two code sets may differ only in the frequency of a single token, and on this basis alone, the classifier can reach 100\% accuracy. However, removing this token and retraining the classifier with $n{-}1$ features would cause a dramatic accuracy drop, indicating that the separability is fragile.
Our robustness metric is designed to quantify how resilient the differences between two code sets are (\reqboxone{R1.2}), i.e., we iteratively remove the most important feature, retrain the classifier, and measure its accuracy. The metric is initially defined as the number of features that can be removed before the accuracy drops below a threshold, e.g., 60\%. However, this threshold would significantly impact the robustness value. Through iterative refinement, we instead define robustness as \textit{the area under the accuracy curve} obtained by progressively removing the most discriminative features. Since our classifier is binary, an accuracy of $x$\% is equivalent to $(100{-}x)$\%. Therefore, we accumulate the absolute deviation from the 50\% accuracy baseline:
\vspace{-8pt}
\begin{equation}\vspace{-4pt}
    \text{Robustness} = \sum_{s=1}^{S} \bigl|\,\text{accuracy}(s) - 0.5\,\bigr|,
    \label{eq:robustness}
\vspace{-3pt}\end{equation}
where $\text{accuracy}(s)$ is the classifier accuracy after the $s$-th most important feature has been removed and $S$ is the total number of feature removal steps ($S \leq 500$). Robustness measures the \emph{persistence} of separability, not its initial level. \textbf{\textit{A large robustness}} means the discriminative signal is present in many features, and the classifier stays meaningfully above random guessing across many removal steps. \textbf{\textit{A small robustness}} means the signal is fragile---this can arise either because accuracy was always near 50\% (the two code sets were never separable), or because accuracy was initially high but collapsed after removing a few features.

\textbf{Concentration.}
While robustness captures how long the classifier remains accurate, it says nothing about how the discriminative signal is structured within the remaining tokens (\reqboxone{R1.3}). Two code sets could be separable due to a single dominant token or the joint effect of multiple tokens. We address this with a complementary metric derived from the entropy ($H(s)$) of the token importance distribution. At each removal step $s$, we compute the entropy of the top-15 (empirically determined) token importance values:
\vspace{-8pt}
\begin{equation}\vspace{-4pt}
  H(s) = -\sum_{i=1}^{15} p_i(s)\,\log_2 p_i(s),
  \quad p_i(s) = {w_i(s)}/{\displaystyle\sum_{j=1}^{15} w_j(s)},
  \label{eq:entropy}
\vspace{-3pt}\end{equation}
where $w_i(s)$ is the importance of the $i$-th ranked token at step $s$. $H(s)$ ranges from $0$ (all importance concentrated in a single token) to $\log_2 15 \approx 3.91$ bits (uniform importance across the 15 tokens).
We define concentration as the integral of the gap between the maximum-entropy ceiling and the observed entropy curve over removal steps, i.e., \textit{the area above the entropy curve}:
\vspace{-8pt}
\begin{equation}\vspace{-4pt}
  \text{Concentration} = \sum_{s=1}^{S/2} \bigl(\log_2 15 - H(s)\bigr).
  \label{eq:concentration}
\end{equation}
Note that the accumulation stops at the first half of steps ($S/2$). This is because, towards the end of feature removal, the remaining features are reduced to just a few and become dominant by default, as no other features remain. The entropy curve and its drop towards the end of feature removal are shown in Fig.~\ref{fig:system}-c2 and explained in Sec.~\ref{sec:ablation_view}.

\textbf{\textit{A high concentration}} indicates that at every removal step, a few tokens always dominate the remaining importance distribution. After the most important token is stripped away, another rises. The two LLMs are therefore distinguished by a \emph{cascade of dominant tokens}---their coding styles differ through a sequence of clear, individually decisive signals.
\textbf{\textit{A low concentration}} indicates that after each removal, the remaining importance distribution is relatively flat. The two LLMs differ in a \emph{diffuse way}---no token jumps out as individually decisive.

\begin{figure}[tb]
  \centering 
  \includegraphics[width=\columnwidth]{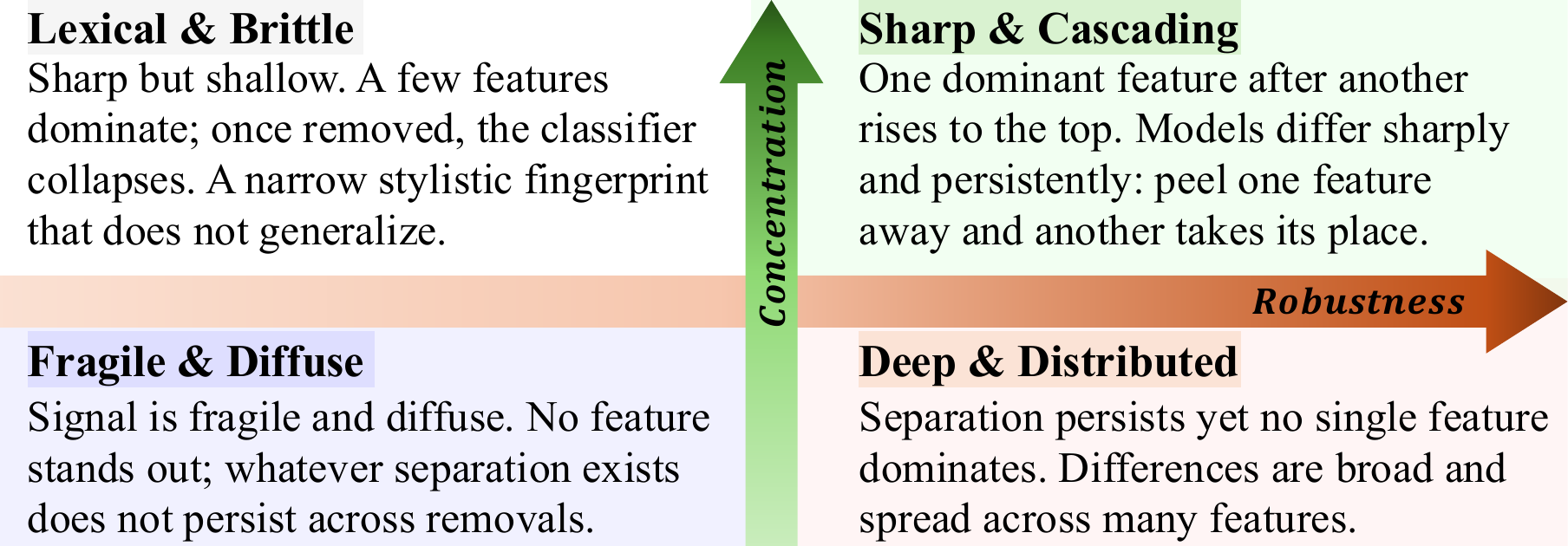}
  \vspace{-0.2in}
  \caption{The four regimes defined by robustness and concentration.}
  \label{fig:robust_concentration}
\end{figure}

All three metrics are needed for a complete picture. Accuracy captures the \emph{initial level} of separability; robustness captures its \emph{persistence} across feature removals; concentration captures the \emph{structure} of the discriminative signal across steps. 
Fig.~\ref{fig:robust_concentration} characterizes the four regimes defined by robustness and concentration, and should be read in conjunction with accuracy to determine whether low robustness reflects genuine low separability or merely a fragile high-accuracy classifier.

\setlength{\belowcaptionskip}{-10pt}
\setlength{\belowcaptionskip}{0pt}

\subsection{Five Code Tokenizers for Multi-Level Analysis}
\label{sec:tokenizer}
Code consists of different semantic levels---comments, functional code, and package usages---each reflecting a distinct aspect of LLM coding behavior. \textit{Analyzing separability at each level independently allows observed differences to be attributed to specific parts of the code (\reqboxthree{R3}).} We introduce five tokenizers, each targeting a different level.

The \textbf{\texttt{complete}} tokenizer extracts all tokens from the entire code without filtering, capturing the full vocabulary including code identifiers, comments, and string literals. It uses a simple regex pattern (\texttt{\textbackslash b[a-zA-Z\_][a-zA-Z0-9\_]*\textbackslash b}) applied directly to the code. This tokenizer provides a comprehensive view of an LLM's vocabulary usage and encompasses all tokens extracted by the other four tokenizers.

The \textbf{\texttt{comment}} tokenizer tokenizes only the inline comments (starting with \texttt{\#}) and docstrings (triple-quoted strings). It uses Python's \texttt{tokenize} module to identify inline comments and abstract syntax tree (AST)~\cite{alfred2007compilers} parsing to extract docstrings from modules, functions, and classes. This tokenizer focuses on an LLM's documentation and explanation style, revealing how LLMs communicate intent and provide context through natural language rather than code.

The \textbf{\texttt{code}} tokenizer is complementary to the \texttt{comment} tokenizer, extracting all tokens from code while excluding comments. This includes code identifiers (variable/function/class names) and string literals within the code (such as those in \texttt{print} statements). It uses the same tools as the \texttt{comment} tokenizer to accurately identify and remove comments/docstrings before applying regex extraction. This tokenizer captures an LLM's programming vocabulary and naming conventions.

The \textbf{\texttt{functional}} tokenizer extracts only identifiers that contribute to the functional logic of the code, producing a strict subset of the \texttt{code} tokenizer's output. It uses a custom AST visitor to traverse the syntax tree and collect identifiers from functional constructs (e.g., variables, functions, classes, etc.) while explicitly skipping non-functional code through predefined rules, such as \texttt{print} statements, \texttt{logging} calls, and display output (e.g., \texttt{tqdm}). This tokenizer isolates an LLM's core algorithmic and computational vocabulary from auxiliary code.

The \textbf{\texttt{package}} tokenizer extracts only the base names of imported packages. It parses import statements (\texttt{import X} or \texttt{from X import Y}) using AST to identify module names, keeping only the base module name (e.g., \texttt{sklearn} from \texttt{sklearn.model\_selection}). It reveals an LLM's library usage patterns and ecosystem preferences.

Our Appendix includes more details on the implementation of the five tokenizers and an example tokenization result from a code snippet.

%% file: tex/6system.tex
\section{Visual Analytics System}
\label{sec:system}
\sysname{} is a visual analytics system comprising three coordinated views that together support a top-down, hypothesis-driven exploration (Fig.~\ref{fig:system}): (a) the \metricview{} provides an overview of pairwise LLM comparisons across tasks and tokenizers, enabling users to identify LLM pairs of interest; (b) the \treeview{} visualizes the trained classifier and supports interactive exploration of discriminative tokens through code examples; (c) the \ablationview{} enables progressive feature removal to assess the robustness and concentration of observed differences. Tight view coordination allows users to navigate fluidly from aggregate metrics down to individual tokens and their code contexts.
\subsection{\metricview{}: Overview of LLM Comparisons}
\label{sec:metric_view}

\textbf{Design Rationale.}
The \metricview{} directly exposes quantitative comparisons across LLM pairs (\reqboxone{R1}). Since we conduct exhaustive pairwise comparisons, a symmetric matrix---where rows and columns represent LLM runs and cell color encodes the metric---is a natural choice. However, the matrix cannot reveal the joint structure of robustness and concentration illustrated in Fig.~\ref{fig:robust_concentration}. We therefore complement it with a scatterplot that plots robustness against concentration, characterizing \textit{how} the separability between different LLM runs is structured.

\textbf{Matrix Mode.} The matrix view (Fig.~\ref{fig:system}a) employs a symmetric matrix layout where both rows and columns represent LLM runs with a naming convention: \texttt{LLM\_T(emperature)\_R(un)}, e.g., \texttt{grok\_T1\_R1} means the first run of the \texttt{grok} model using temperature 1. Each cell encodes a metric value through color intensity using a sequential color scale, where darker shades indicate higher values. Since the matrix is symmetric, the upper-right ($\urblacktriangle$) and lower-left ($\llblacktriangle$) triangles can encode different metrics, selectable via control widgets at the top left. This view presents metrics for one tokenizer at a time; users can switch among the five tokenizers via a dropdown widget in the view header.

\setlength{\belowcaptionskip}{-8pt}
\begin{figure}[tb]
  \centering 
  \includegraphics[width=\columnwidth]{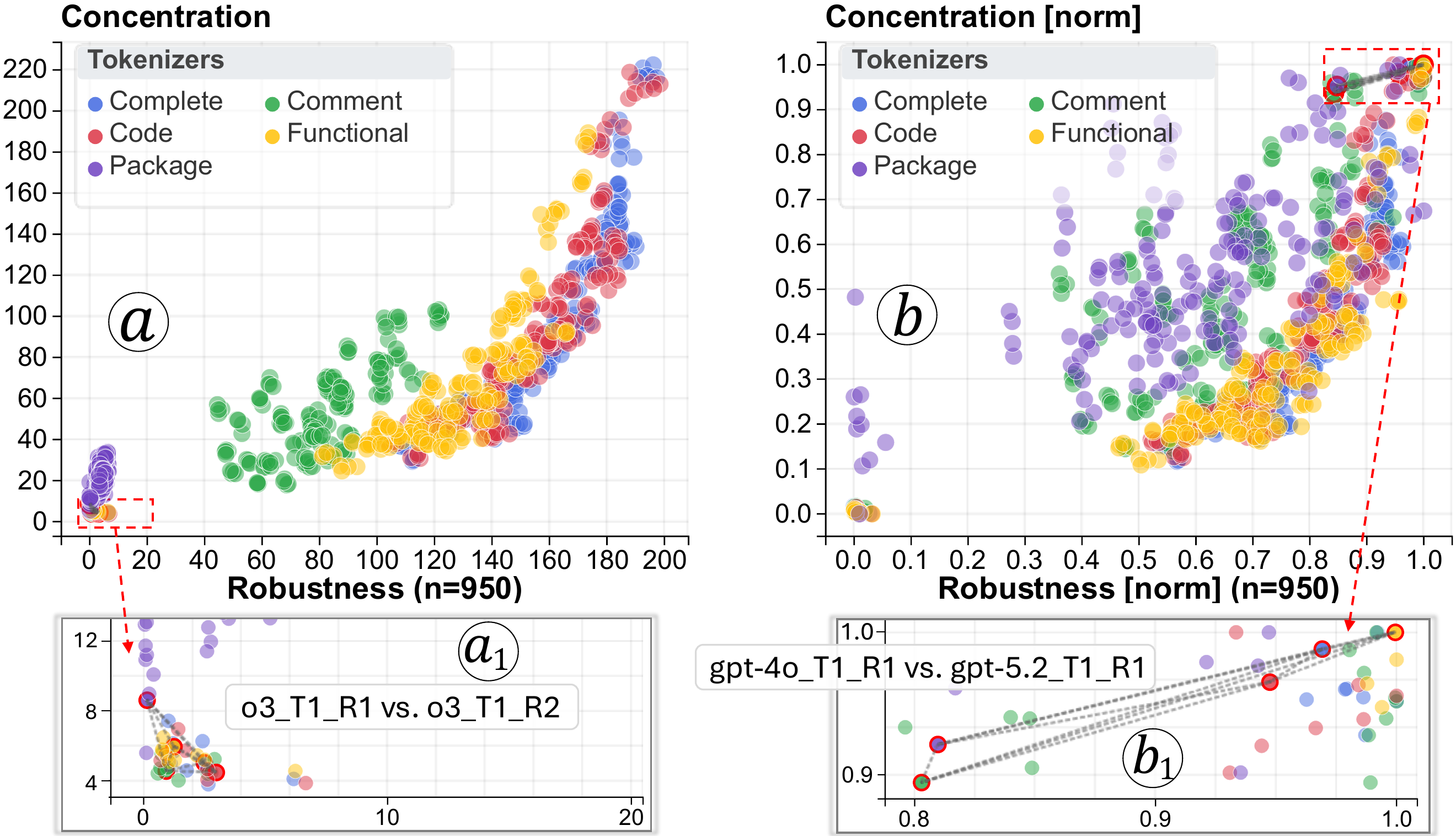}
  \vspace{-0.2in}
  \caption{The scatterplot realizes the design in Fig.~\ref{fig:robust_concentration}: (a) raw values; (b) normalized values. Each point represents a pair of LLM runs. Comparing 20 LLM runs across 5 tokenizers results in 950 ($20{\times}19/2{\times}5$) pairs/points.}
  \label{fig:scatterplot}
\end{figure}
\setlength{\belowcaptionskip}{0pt}

\textbf{Scatterplot Mode.} Each point in the scatterplot (Fig.~\ref{fig:scatterplot}a) represents a pair of LLM runs (i.e., a cell in the matrix mode). The x- and y-axes encode robustness and concentration, respectively. This mode presents five tokenization levels concurrently, and the points are color-coded by their tokenizer type. When all tokenizers are shown, metric values vary considerably across tokenizers. We therefore enable the points to be normalized within each set (Fig.~\ref{fig:scatterplot}b), so that a pair's separability can be compared across tokenizers relative to other pairs (\reqboxthree{R3.2}). For example, in Fig.~\ref{fig:scatterplot}-b1, the separability between \texttt{gpt-4o\_T1\_R1} and \texttt{gpt-5.2\_T1\_R1} is always ``\sharpcascading{Sharp \& Cascading}'' (Fig.~\ref{fig:robust_concentration}, top-right) across all five tokenizers. Dashed lines are drawn to connect points representing tokenizers of the same LLM pair for easy identification.

\textbf{Interactions.}
Users can toggle between the matrix and scatterplot from the view title (see associated video). Both modes support cell or point selection through clicking, which triggers coordinated updates across views, i.e., broadcasting the selection to downstream views. 

Since LLM coding behavior can vary substantially across ML tasks, a task selection dropdown allows users to focus on individual tasks (see details of the 22 tasks in Appendix). To support a holistic comparison, we also provide an ``\textbf{All Tasks}'' mode that trains classifiers on all 22 tasks jointly, separating two code sets of $22{\times}1,000$ samples each.

\subsection{\treeview{}: Interactive Token Exploration}
\label{sec:tree_view}
\textbf{Design Rationale.}
While the \metricview{} answers \textit{\textbf{which}} LLM pairs differ, the \treeview{} explains \textit{\textbf{how}} they differ by visualizing the trained classifier structure. Decision trees are inherently interpretable: each internal node denotes a discriminative feature and its splitting threshold, while leaf nodes indicate classification outcomes. By directly visualizing the tree structure, users can understand the hierarchical decision logic that separates two code sets (\reqboxtwo{R2.1}).

\textbf{Decision Tree Visualization.}
We visualize the tree with a node-link diagram for its intuitive clarity (Fig.~\ref{fig:system}-b1). The encoding is as follows:
\begin{itemize}[leftmargin=0.3cm, topsep=0pt, partopsep=0pt, labelsep=0.1cm, itemsep=-0.1cm]
    \item \textit{Node label}: Shows the splitting token and threshold for internal nodes (e.g., \texttt{try$\leq$0.5}: samples with no \texttt{try} go to the Yes branch) and the classified set for leaf nodes; sample counts are shown in both.
    \item \textit{Node color}: The diverging color bar at the bottom encodes the sample distribution in a node---blue for Set 1, green for Set 2. The gray and white background indicate internal and leaf nodes, respectively.
    \item \textit{Edge width}: Width is proportional to code sample count.
\end{itemize}

\textbf{Feature Importance Panel.}
A horizontal bar chart (Fig.~\ref{fig:system}-b2) next to the tree visualization shows the top-15 most important tokens, ranked by their contribution to the separability (\reqboxtwo{R2.2}). Each bar encodes a token's importance score---its cumulative Gini impurity decrease across all nodes, ranging from 0 to 1. This panel serves dual purposes: (1) providing a ranked summary of discriminative tokens without requiring tree traversal, and (2) enabling direct token selection for code context retrieval or feature ablation experiments (see Interactions below).

\textbf{Interactions.}
This view supports three primary interactions:

\textit{Tree Navigation:} Users can expand (\includegraphics[height=0.8em]{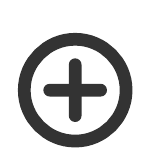}) or collapse (\includegraphics[height=0.8em]{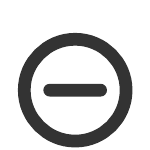}) subtrees by clicking individual nodes, allowing progressive disclosure of tree complexity. The tree is initially rendered with five levels---sufficient for most analyses---with deeper subtrees expandable on demand. Panning and zooming further support exploration of large trees. 

\setlength{\belowcaptionskip}{-11pt}
\begin{figure*}[tb]
  \centering 
  \includegraphics[width=\textwidth]{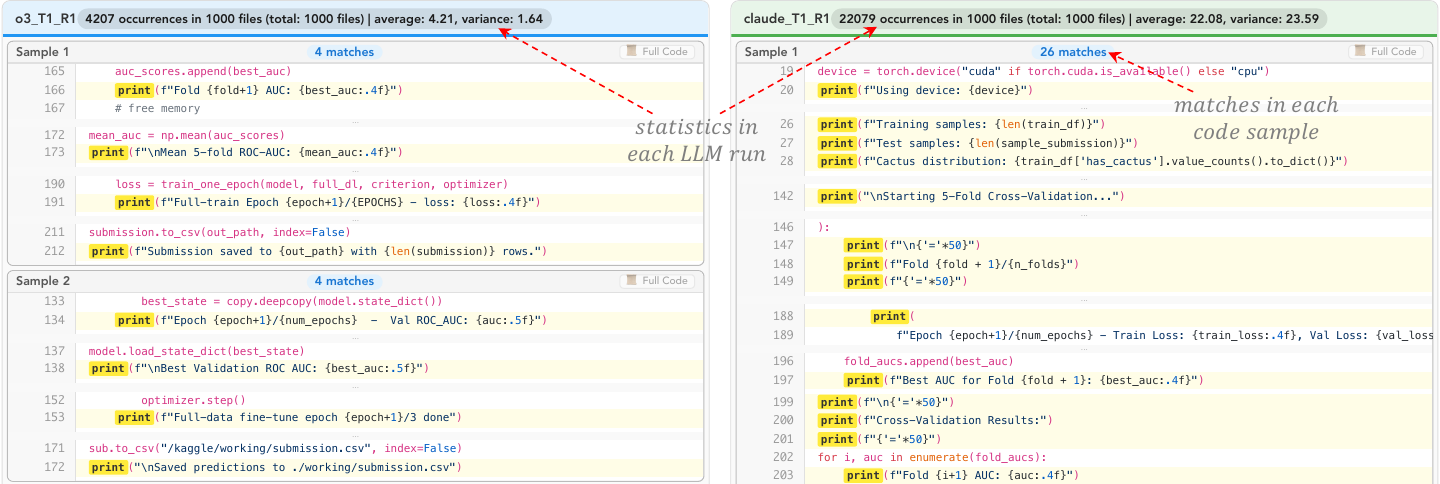}
  \vspace{-0.22in}
  \caption{The side-by-side comparison of code samples containing the token \texttt{print} from the two code sets (left: \texttt{o3\_T1\_R1}, right: \texttt{claude\_T1\_R1}). The statistics show that, on average, \texttt{claude} uses \texttt{print} 22.08 times in each of its code samples. In contrast, \texttt{o3} uses it much less frequently.}
  \label{fig:code}
\end{figure*}
\setlength{\belowcaptionskip}{0pt}

\textit{Token-to-Code Linking:} Clicking any token in the Feature Importance Panel (Fig.~\ref{fig:system}-b2) triggers a \codeview{} (Fig.~\ref{fig:code}) that retrieves code snippets containing the selected token. The view shows side-by-side comparisons of both LLM runs' code samples, with the queried token highlighted. Occurrence statistics (count, average, variance) are displayed for each LLM run, enabling users to ground abstract feature importance scores in concrete code examples. A search bar (Fig.~\ref{fig:system}-b2) further enables users to search for a specific token, supporting hypothesis-driven exploration, e.g., searching \texttt{try} or \texttt{except} to test whether the two LLMs differ in exception handling.

\textit{Feature Ablation:} Clicking the feature importance bars in Fig.~\ref{fig:system}-b2 initiates dynamic feature removal. The selected token is added to an exclusion list, and the system retrains the classifier without tokens in the list. The tree view updates in real-time with the new tree structure, allowing users to observe how classification logic adapts when discriminative tokens are removed (\reqboxtwo{R2.3}). Excluded tokens are tracked in the \ablationview{} (Sec.~\ref{sec:ablation_view}). This interaction supports ``what-if'' analysis: users can test whether observed differences are robust or fragile. Feature ablation can also uncover features hidden by an equally dominant one: if two features each perfectly separate the two code sets, the tree uses only the first and assigns it an importance of 1, leaving the second invisible in the panel. Removing the first reveals the second.

\subsection{\ablationview{}: Progressive Feature Removal}
\label{sec:ablation_view}
\textbf{Design Rationale.} The \ablationview{} (Fig.~\ref{fig:system}c) operationalizes the robustness and concentration metrics through progressive feature removal at scale. While accuracy captures the initial separability level, it does not reveal the depth or structure. The \treeview{} supports single-step removal, but computing robustness and concentration requires removing hundreds of features in sequence, a process that would be prohibitively tedious to perform manually. The \ablationview{} automates this and visualizes the metrics with cumulative details.

\textbf{Tri-Axis Encoding.}
The view employs a tri-axis combination chart (Fig.~\ref{fig:system}c) whose x-axis represents the token removal sequence. Three y-axes encode complementary metrics:
\begin{itemize}[leftmargin=*, topsep=0pt, partopsep=0pt, labelsep=0.1cm, itemsep=-0.1cm]
    \item \textit{Left y-axis (bars):} Each bar (Fig.~\ref{fig:system}-c1) encodes the importance of the token removed (i.e., the top-1) at that step, representing the upper bound of a token's importance. The importance typically decreases as more tokens are removed, but rises toward the end when only a few tokens remain and each dominates by default. Concentration is therefore computed over only the first half of removals (Fig.~\ref{fig:system}-c2). 
    \item \textit{Right primary y-axis (orange):} An orange curve encodes classification accuracy after each removal, showing how separability degrades as discriminative tokens are progressively eliminated. A dashed orange line marks the 50\% baseline, and the orange area is the robustness.
    \item \textit{Right secondary y-axis (green):} A green curve encodes the entropy of the top-15 feature importances after each removal, reflecting how concentrated the discriminative signal is. A dashed green line marks the maximum entropy ($\log_2 15$), and the green area is the concentration.
\end{itemize}

\textbf{Interactions.}
The view supports manual removal, automated batch removal, and retrospective inspection of any removal step.

\textit{Manual Removal:} Tokens removed via the \treeview{}'s Feature Importance Panel are immediately reflected here. Each removal appends a new bar, accuracy point, and entropy point to the chart, with animated transitions extending the visualization rightward. Users can thus observe whether each removed feature is critical or redundant.

\textit{Automated Batch Removal:} A control panel (Fig.~\ref{fig:system}-c3) lets users specify a target removal count (1--500 features) and trigger batch removal via a ``Start'' button. The system iteratively removes the currently most important feature, retrains the classifier, and updates the chart until the target count is reached. A progress overlay (e.g., ``Removing feature 47/500\ldots'') keeps users informed, with the option to ``Stop'' or ``Reset'' at any time. This mode is essential for cases where users want to remove many tokens, in which manual removal is impractical.

\textit{Time Travel:} Clicking any bar restores the decision tree to the state at that removal step. For example, clicking the 10th bar loads the tree after the first 9 features have been removed. Users can identify where accuracy dropped sharply and examine the corresponding tree to understand which remaining features carry more discriminative power.

\textit{Zoom and Pan.}
With hundreds of removal steps, the chart becomes compressed; users can zoom into specific regions (Fig.~\ref{fig:system}-c1) to locate significant accuracy oscillation that may be invisible in the full view.

\textit{Checkpoint Caching.}
Since users often revisit the same pair of LLM runs across sessions, the system automatically caches each removal step keyed by (LLM run pair, task, tokenizer). Because \textit{exhaustive} feature removal has already been performed offline when computing robustness and concentration, these precomputed results allow the system to restore any removal step instantly when a pair is selected.

%% file: tex/7casestudy.tex
\section{Case Study and Actionable Insights}
\label{sec:casestudy}
This section compares code sets generated by different LLMs, using different tokenizers, at different LLM temperatures, and with different prompts. Through these comparisons, we reveal fundamental differences between LLMs' coding styles and demonstrate how the comparative insights help users better control LLMs' coding behavior.

\setlength{\belowcaptionskip}{-6pt}
\begin{figure*}[tb]
  \centering 
  \includegraphics[width=\textwidth]{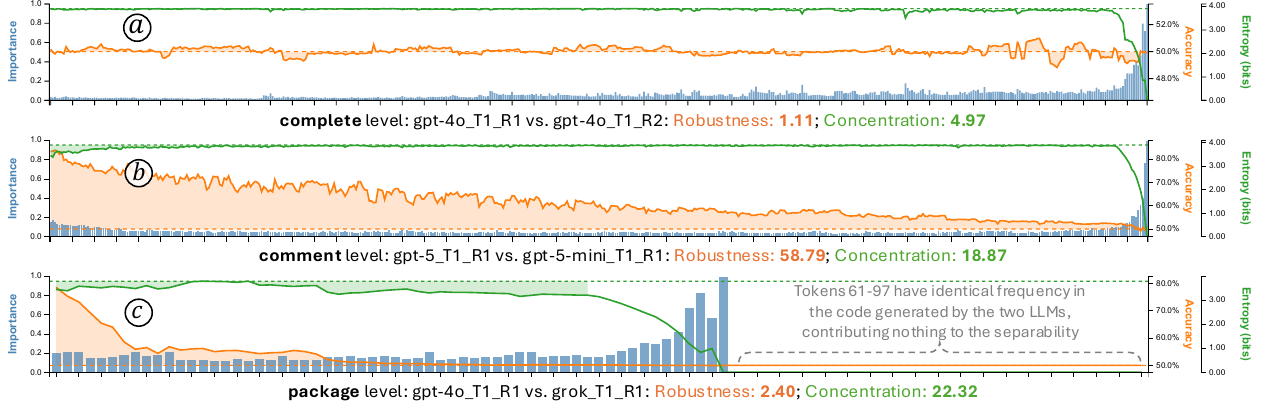}
  \vspace{-0.2in}
  \caption{The \ablationview{} when comparing different pairs of LLM runs in different tokenization levels.}
  \label{fig:ablation}
\end{figure*}
\setlength{\belowcaptionskip}{0pt}

\subsection{Comparing Different LLMs}
\label{sec:all-tasks}
To gain an overall impression, we first compare the 10 LLMs (Sec.~\ref{sec:codegeneration}) using the ``All Tasks'' mode with the \texttt{complete} tokenizer. The 20 LLM runs appear as 20 rows/columns in the matrix of Fig.~\ref{fig:system}a.
Each cell represents a comparison between two code sets, each with $22{,}000$ code samples ($22\ \text{tasks} {\times} 1{,}000\ \text{code samples per LLM run}$).

The matrix shows that code from the \emph{\textbf{same}} LLM's two independent runs is not separable: light blue cells near the diagonal in the upper-right triangle indicate $\approx$50\% accuracy, and light orange cells near the diagonal in the lower-left triangle confirm low robustness. We click a representative cell, \texttt{gpt-4o\_T1\_R1} vs.\ \texttt{gpt-4o\_T1\_R2} (Fig.~\ref{fig:system}-a1), and examine it in the \ablationview{}. As shown in Fig.~\ref{fig:ablation}a, the accuracy curve is always around 50\%, reflecting low separability and a low robustness value of 1.11. The entropy curve is always around $\log_{2}15$ (the max value), indicating that no dominant features contribute to the separation, resulting in a very low concentration value of 4.97.

In contrast, for any two runs from \emph{\textbf{different}} LLMs, the separability is very high, as reflected by the dark blue cell color in Fig.~\ref{fig:system}a. Switching to the scatterplot mode, we select a point at the far top-right corner with ``\sharpcascading{Sharp \& Cascading}'' separability, representing \texttt{gpt-4o\_T1\_R1} vs.\ \texttt{gpt-5.2\_T1\_R1} (Fig.~\ref{fig:system}-a2). The corresponding \ablationview{} is shown in Fig.~\ref{fig:system}c, in which the orange and green areas are very large (robustness: $192.49$, concentration: $220.97$), echoing the persistent separability and concentrated discriminative signal.

We then use the \treeview{} to identify which tokens drive this separation. As shown in Fig.~\ref{fig:system}-b2, the separability is dominated by token \texttt{exist\_ok}, the root splitting feature of the decision tree (Fig.~\ref{fig:system}-b1). When \texttt{exist\_ok} appears fewer than 0.5 times (the \texttt{Yes} branch), the code is classified as Set 1 (\texttt{gpt-4o\_T1\_R1}); otherwise as Set 2 (\texttt{gpt-5.2\_T1\_R1}). Based on this single token's frequency, the tree can almost correctly separate the $35{,}200$ code samples from the two sets, i.e., $2{\times}22{,}000{\times}80\%$ (the training data is 80\% of all code samples), as the two second-level tree nodes have almost pure blue or green color. 

Inspecting token occurrences via the \codeview{} (Fig.~\ref{fig:code}), \texttt{exist\_ok} appears exclusively in two patterns:
\begin{lstlisting}[
  aboveskip=2pt,
  belowskip=2pt,
  xleftmargin=3em,
  literate=
    {os}{{\textcolor{teal!80!black}{os}}}2
    {makedirs}{{\textcolor{violet!80!black}{makedirs}}}8
    {mkdir}{{\textcolor{violet!80!black}{mkdir}}}5
    {exist_ok}{{\setlength\fboxsep{1pt}\colorbox{yellow}{\textbf{\textcolor{orange!70!black}{exist\_ok}}}}}8
    {parents}{{\textcolor{orange!70!black}{parents}}}7,
]
os.makedirs(WORKING_DIR, exist_ok=True)
WORKING_DIR.mkdir(parents=True, exist_ok=True)
\end{lstlisting}
Among the $22{,}000$ code samples generated by \texttt{gpt-4o\_T1\_R1}, only 20 contain \texttt{exist\_ok}; for \texttt{gpt-5.2\_T1\_R1}, the count is 21,922. This stark contrast makes \texttt{exist\_ok} a fingerprint of \texttt{gpt-5.2}. As long as it occurs, we can confidently conclude that the code is from \texttt{gpt-5.2}.

The code context of this token also leads us to hypothesize that \textit{\texttt{gpt-5.2} is more cautious about data writing, i.e., using \texttt{makedirs} or \texttt{mkdir} to proactively create the directory in case it does not exist}. To verify this, we use the search bar in Fig.~\ref{fig:system}-b2 to retrieve the occurrence counts for \texttt{makedirs} and \texttt{mkdir}. As shown in Tab.~\ref{tbl:mkdir}, \texttt{gpt-5.2} uses these two tokens in $21{,}922$ out of its $22{,}000$ code samples, while \texttt{gpt-4o} does so only in $25$ out of its $22{,}000$ samples. This stark frequency contrast convincingly confirms our hypothesis. The case also illustrates \sysname{}'s hypothesis-driven workflow: a discriminative token (\texttt{exist\_ok}) surfaces a behavioral signal, and the search bar lets users immediately pursue the underlying cause, confirming a systematic divergence in how the two LLMs handle directory creation.

\setlength{\intextsep}{12pt}
\begin{table}[thb]\small
\caption{The frequency of \texttt{makedirs} and \texttt{mkdir} in two code sets.}
\label{tbl:mkdir}
\vspace{-0.13in}
\centering
\begin{tabular}{|l|ll|ll|}
\hline
         & \multicolumn{2}{l|}{\texttt{gpt-4o\_T1\_R1}}               & \multicolumn{2}{l|}{\texttt{gpt-5.2\_T1\_R1}}     \\ \hline
         & \multicolumn{1}{l|}{\# occurrences} & \# files & \multicolumn{1}{l|}{\# occurrences} & \# files \\ \hline
\texttt{\textbf{makedirs}} & \multicolumn{1}{l|}{33}                   & 25    & \multicolumn{1}{l|}{16,149}       & 16,138 \\ \hline
\texttt{\textbf{mkdir}}    & \multicolumn{1}{l|}{0}                    & 0     & \multicolumn{1}{l|}{5,791}        & 5,784  \\ \hline
total    & \multicolumn{1}{l|}{33}                   & \textbf{25}/22,000    & \multicolumn{1}{l|}{21,940}       & \textbf{21,922}/22,000 \\ \hline
\end{tabular}
\vspace{-0.2cm}
\end{table}
\setlength{\intextsep}{12pt} 

To verify whether the two code sets only differ in \texttt{exist\_ok}, we remove it and retrain the tree with the remaining 499 features. As shown in Fig.~\ref{fig:system}-c1, the token \texttt{tr\_idx} becomes the most important feature, with an importance of 0.9122. This cascading dominance reflects the concentrated discriminative signal, echoing the high concentration. 

\textbf{Per-Task Comparison.} 
We have also compared LLM runs using the code of individual tasks, i.e., differentiating $2{\times}1{,}000$ code samples. \textit{The general trend is consistent with the ``All Tasks'' mode} where we differentiate $2{\times}22{,}000$ code samples (the more comprehensive comparison). However, the most discriminative tokens vary across individual tasks, as each task has a particular problem type and data modality. For example, the token \texttt{torchvision} appears more prominently in image-related tasks but rarely appears in tasks of tabular data. As the per-task comparison does not introduce extra insights beyond the ``All Tasks'' mode, we include its details in our Appendix.

\subsection{Comparing LLMs using Different Tokenizers}
Next, we switch to different tokenizers to compare the findings with those from the \texttt{complete} tokenizer in Sec.~\ref{sec:all-tasks}.

The \texttt{functional} and \texttt{code} tokenizers behave very similarly to the \texttt{complete} tokenizer: code from the same LLM is inseparable, while code from different LLMs is highly separable. The corresponding three sets of 190 ($20{\times}19/2$) LLM run pairs also follow similar distributions in the scatterplot of Fig.~\ref{fig:scatterplot}a (the yellow, red, and blue points). There is a slight shift toward the right (more robust) from the yellow (\texttt{functional}), to the red (\texttt{code}), to the blue (\texttt{complete}) clusters. We believe this is due to the inclusive relationship among the token sets produced by the three tokenizers, i.e., \texttt{functional}$\subset$\texttt{code}$\subset$\texttt{complete}.
All LLM pairs in the three tokenization levels have robust separability, while the concentration of discriminative signals varies across pairs. Some LLM pairs, e.g., \texttt{o3} and \texttt{claude}, can be easily separated by a single token (\texttt{print} in Fig.~\ref{fig:code}), i.e., the separability is ``\sharpcascading{Sharp \& Cascading}'', while others, e.g., \texttt{gpt-5} and \texttt{gpt-5-mini}, need multiple tokens to work collaboratively, i.e., the ``\deepdistributed{Deep \& Distributed}'' separability in Fig.~\ref{fig:robust_concentration}. The LLM pairs with the latter type of separability often share similar LLM versions, indicating that \textit{although the separability between them is still robust, the discriminative signal is not sharp.}

\begin{figure}[tb]
  \centering 
  \includegraphics[width=\columnwidth]{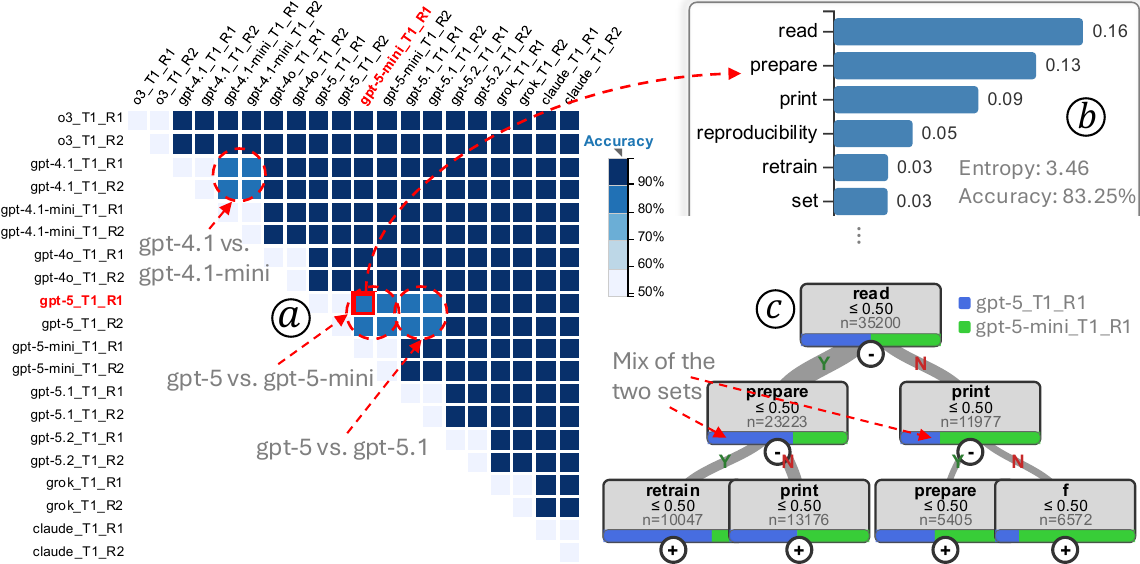}
  \vspace{-0.2in}
  \caption{(a) The \texttt{comment} level accuracy matrix. (b) Diffuse feature importance (i.e., high entropy). (c) Decision tree with mixed sets of code.}
  \label{fig:comments}
\end{figure}

For the \texttt{comment} tokenizer, we notice that the accuracies for pairs (1) \texttt{gpt-4.1} vs.\ \texttt{gpt-4.1-mini}, (2) \texttt{gpt-5} vs.\ \texttt{gpt-5-mini}, and (3) \texttt{gpt-5} vs.\ \texttt{gpt-5.1} are relatively low, as reflected by the three lighter blue blocks in Fig.~\ref{fig:comments}a. These LLM pairs show more similar commenting styles, and we believe their similarity was driven by their similar LLM versions.
From them, we randomly select an LLM run pair to examine its details, i.e., \texttt{gpt-5\_T1\_R1} vs.\ \texttt{gpt-5-mini\_T1\_R1}. As shown by the top-6 important features in Fig.~\ref{fig:comments}b, the discriminative signal is not dominated by any single feature. The corresponding tree visualization in Fig.~\ref{fig:comments}c echoes this. The root node uses the token \texttt{read} to split the two code sets. However, after this splitting, the two second-level tree nodes still show a significant mix of code samples (the mix of blue and green bars), reflecting the weak discriminative power of \texttt{read}.
The corresponding \ablationview{} is shown in Fig.~\ref{fig:ablation}b, where the orange and green areas reflect the relatively lower robustness and much lower concentration values.
Extending the analysis to more pairs, we found that the \texttt{comment} level often shows relatively lower robustness and concentration compared to the \texttt{code} level, as shown by the green points in Fig.~\ref{fig:scatterplot}a, indicating there are fewer dominant tokens in the comments. This is expected, as comments are in natural language and share paraphrasable vocabularies, spreading the discriminative signal across many tokens. In contrast, code must obey strict syntax, so each LLM's preferences can manifest as individually decisive token choices. \textit{In short, the discriminative power of comment tokens is weaker than that of code tokens, as code needs to follow much stricter syntax}.

For the \texttt{package} level, the accuracy, robustness, and concentration values are generally lower. This is because the number of unique tokens (i.e., unique packages) is much smaller, resulting in much shorter feature vectors when training the decision trees. While exploring the \metricview{} (screenshots in the Appendix), we notice that the classifier separating \texttt{gpt-4o\_T1\_R1} and \texttt{grok\_T1\_R1} shows noticeably lower accuracy. Fig.~\ref{fig:ablation}c shows the \ablationview{} when comparing this pair. There are only 97 unique features from the two code sets (other levels have 500 features). Among them, 37 have the same frequency (features 61--97 in Fig.~\ref{fig:ablation}c), contributing nothing to the separability. This further suggests that \texttt{gpt-4o} and \texttt{grok-4} share similar package preferences. \textit{To conclude, the \texttt{package} tokenizer produces the most ``\fragilediffuse{Fragile \& Diffuse}'' separations (Fig.~\ref{fig:scatterplot}a): its feature space is small (fewer unique packages), and capable LLMs tend to converge on similar optimal libraries for the same tasks, leaving little stylistic divergence.} 

Note that for the two runs generated by the same LLM, their separability is always ``\fragilediffuse{Fragile \& Diffuse}'' across all five tokenizers, because these runs are indistinguishable. Fig.~\ref{fig:scatterplot}-a1 shows one example, i.e., \texttt{o3\_T1\_R1} vs.\ \texttt{o3\_T1\_R2}. The five points representing the five tokenizers for the pair are connected by dashed lines for easy identification. 

\textbf{Regime Coverage.} We have seen separability types covering three regimes explained in Fig.~\ref{fig:robust_concentration}. Only the ``\lexicalbrittle{Lexical \& Brittle}'' one is not covered. This regime works as a sanity check on the generalizability of our trained classifiers. Specifically, the feature importance can only be computed from the training data, but the classifier accuracy is derived from the test data. A ``\lexicalbrittle{Lexical \& Brittle}'' separability would mean that dominant features exist (high concentration) during training and significantly reduce the impurity when separating the two code sets, yet the test accuracy remains around 50\% (low robustness). This contradiction leads to the conclusion that the trained classifier does not generalize well to the test data. The absence of LLM run pairs in this regime suggests consistency between the training and test data distributions and supports the generalizability of our trained classifiers.

\setlength{\belowcaptionskip}{-10pt}
\setlength{\belowcaptionskip}{0pt}

\subsection{Comparing LLMs with Different Temperatures}
To investigate the impact of LLMs' temperature on their code generation behavior, we varied the temperature (0.3, 0.5, 0.7, and 1) of two LLMs, \texttt{grok} and \texttt{claude}, and generated additional code following our dual-run mechanism (Sec.~\ref{sec:codegeneration}). This code corpus contains $2\ (LLMs) {\times}\allowbreak 22\ (tasks) {\times}\allowbreak 2\ (runs) {\times}\allowbreak 1{,}000\ (code) {\times}\allowbreak 4\ (temperatures) {=} 352{,}000$ code samples. We focus on the \texttt{complete} tokenizer and use the ``All Tasks'' mode, as they provide the most comprehensive characterization.

\begin{figure}[tb]
  \centering 
  \includegraphics[width=\columnwidth]{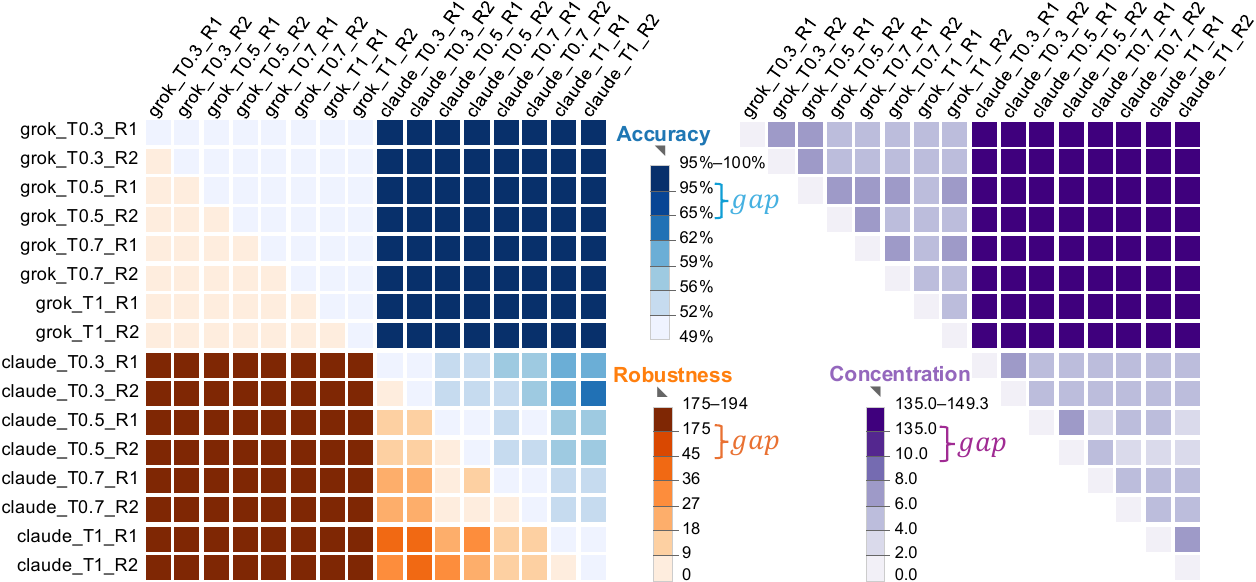}
  \vspace{-0.2in}
  \caption{Comparing \texttt{grok} and \texttt{claude} with different temperatures.}
  \label{fig:temp}
\end{figure}

Fig.~\ref{fig:temp} shows the \metricview{}. Note that we adjusted the color mapping to emphasize value differences in the lower range (see the legends). As expected, accuracy is high when the compared code sets are from different LLMs. When comparing the same LLM with different temperatures, \texttt{grok} shows no obvious difference, as the accuracy values are always around 50\%. In contrast, \texttt{claude} shows differences that become more pronounced as the temperature gap widens (i.e., between 0.3 and 1). However, the difference is not dramatic, as the accuracy is always below 65\% according to the color legend. The robustness metric follows a similar pattern to accuracy. The concentration metric, however, shows no notable variation across temperatures and remains very small, indicating no single token dominates the separability.

We further explored individual tokens using the \treeview{} and \ablationview{}, but no dominant discriminative features emerged, corroborating the low concentration values above. Taken together, while temperature has limited impact on coding style, \texttt{claude} exhibits richer temperature-sensitive behavior than \texttt{grok}, suggesting that temperature is a more effective lever for steering \texttt{claude}.

\setlength{\belowcaptionskip}{-8pt}
\setlength{\belowcaptionskip}{0pt}

\subsection{Comparing LLMs with Different Prompts}
\label{sec:casestudy_prompt}
\begin{figure}[b]
  \centering 
  \includegraphics[width=\columnwidth]{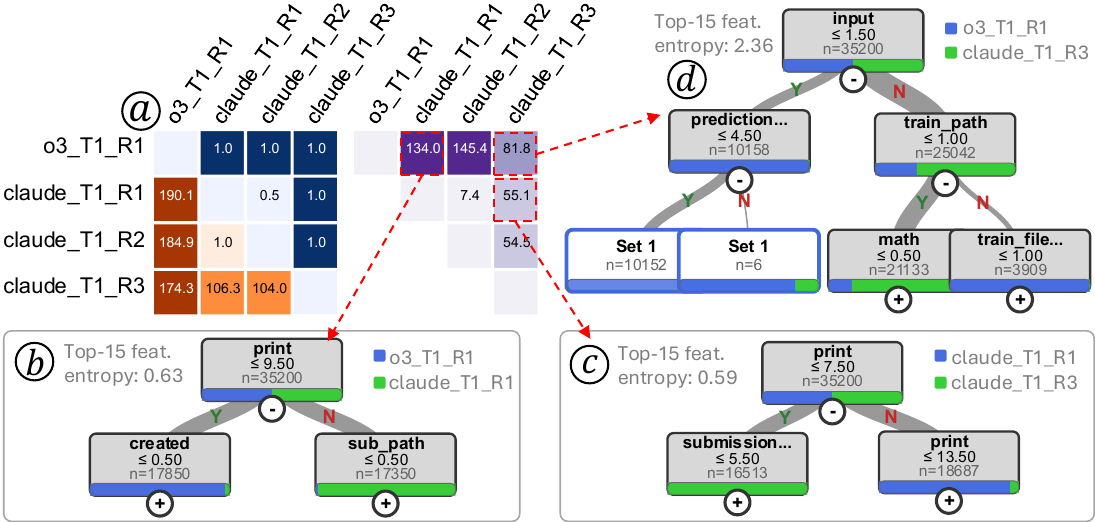}
  \vspace{-0.2in}
  \caption{Comparing LLMs when using different coding prompts.}
  \label{fig:prompt}
\end{figure}

Comparing \texttt{o3} with \texttt{claude}, we found that token \texttt{print} dominates the separation. As shown in the corresponding decision tree (Fig.~\ref{fig:prompt}b), when \texttt{print} occurs less than 9.5 times, the code is more likely to be from \texttt{o3}, indicating that \texttt{claude} is much more verbose than \texttt{o3} (Fig.~\ref{fig:code} echoes this). In practice, users may need to use \texttt{claude} but control its verbosity. This motivates us to steer \texttt{claude}'s coding behavior. Specifically, we added the following to the prompt of \texttt{claude} and generated a new run of code. Following our naming convention, this new run is \texttt{claude\_T1\_R3}.

\textit{``When writing the code, keep output minimal. Avoid verbose prints, debug messages, chatty logs, or explanatory console text. Only include print/log statements if they are strictly necessary for the program's functionality (e.g., required output format).''}

Comparing \texttt{claude\_T1\_R3} to \texttt{claude\_T1\_R1}, the classifier reaches 100\% accuracy (see values inside the cells of Fig.~\ref{fig:prompt}a) and \texttt{print} is the dominant feature (Fig.~\ref{fig:prompt}c). When \texttt{print} occurs less than 7.5 times, the code is more likely to be from \texttt{claude\_T1\_R3}, indicating that the prompt successfully suppressed the verbosity of \texttt{claude}. This new set of code is also sufficiently separable from \texttt{claude\_T1\_R1} even after removing \texttt{print}, with robustness of 106.3 and concentration of 55.1. For reference, when comparing \texttt{claude\_T1\_R2} to \texttt{claude\_T1\_R1}, these two values are only 1.0 and 7.4, respectively (Fig.~\ref{fig:prompt}a). The much higher robustness and concentration reflect that the new prompt not only changes the verbosity but also other intricate coding behaviors.

When comparing \texttt{claude\_T1\_R3} to \texttt{o3\_T1\_R1}, \texttt{print} is no longer the dominant feature (Fig.~\ref{fig:prompt}d), because both sets are less verbose. In fact, \texttt{print} was never used in the top three levels of the tree as a splitting feature. The entropy of the top-15 important features is also much higher (2.36 bits), indicating no single feature dominates the separability. Notably, \texttt{claude\_T1\_R3} is more distant from \texttt{o3\_T1\_R1} than from \texttt{claude\_T1\_R1}, because the robustness ($174.3{>}106.3$) and concentration ($81.8{>}55.1$) are both larger (Fig.~\ref{fig:prompt}a). This implies that reducing verbosity in \texttt{claude} does not simply bring it closer to \texttt{o3} but instead exposes deeper stylistic differences between the two LLMs. The classifier accuracy alone cannot reflect this relative distance, as it is always very close to 100\% (the dark blue cells in Fig.~\ref{fig:prompt}a).

%% file: tex/8feedback.tex
\section{Feedback from Target Users}
\label{sec:feedback}
We further evaluate \sysname{} with ML scientists and engineers---our target users---in two complementary settings: expert think-alouds for \textit{in-depth} design feedback, and a large-scale practitioner seminar for \textit{broad} signal.

\textbf{Think-Aloud + Guided Explorations.}
In the first setting, we conducted think-aloud sessions with the three experts ($E_1$--$E_3$) who contributed to our requirement analysis. We walked them through \sysname{} and encouraged free exploration. The sessions totaled over six hours across two weeks. All experts quickly grasped the mechanics of \sysname{} and engaged enthusiastically.
$E_1$, who proposed \textit{area under the accuracy curve} for robustness, expressed strong appreciation for seeing the idea realized, noting that this formulation is hyperparameter-free. $E_2$, whose work centers on feature engineering, confirmed that progressive feature ablation is an effective lens for studying separability. He was particularly impressed by the prompt engineering study (Sec.~\ref{sec:casestudy_prompt}), calling it a practical method for steering LLMs through targeted prompting. $E_3$ offered a more theoretical perspective, confirming the soundness of token-frequency analysis and suggesting that squaring the deviation from 50\% in Eq.~\ref{eq:robustness} could give greater weight to steps with high accuracy. The three experts consistently recognized the unique value of the system. As a strong indication of their interest, all offered to contribute code from their own projects for future comparison. $E_1$ was also excited about extending the framework to compare LLM-generated and human-written code---a direction that could reveal fundamental differences in coding between humans and LLMs.

\textbf{Seminar + Feedback Survey.}
In the second setting, we gave a one-hour seminar to 40+ full-time ML scientists. The seminar began with a structured overview of \sysname{}, followed by a live demonstration using real LLM comparison tasks, with participants encouraged to ask questions freely. Afterward, 35 of them voluntarily completed an anonymized 20-question survey, covering demographics, self-assessed understanding of \sysname{}, perceived utility of each component, and overall impressions. Fig.~\ref{fig:survey} shows the response distributions across 10 key questions (see full responses in Appendix). Overall, participants demonstrated strong appreciation for the system and expressed positive assessments of \sysname{}.
The participants also offered valuable comments. One senior scientist connected our frequency-based analysis to TF-IDF~\cite{sparck1972statistical}, suggesting that token frequency should be normalized by code length to account for verbosity. Two scientists noted that our regex-based tokenizer differs from LLM tokenizers~\cite{sennrich2016neural,wu2016google} that may split a single word into multiple tokens (e.g., from \texttt{tokenization} to \texttt{token} and \texttt{ization}), whereas ours never breaks a word, preserving word-level semantics. Multiple scientists also commented that humans are needed to connect code tokens to LLMs' personas, e.g., from the frequency of \texttt{print} to \texttt{verbose}. This observation highlights a meaningful role for \sysname{}: it serves as the bridge between raw code tokens and human-interpretable LLM personas, making the implicit explicit.

%% file: tex/9discussion.tex
\section{Discussion, Limitations, and Future Work}
\label{sec:discussion}
While we focus on Python code and ML tasks, the approach generalizes readily to other programming languages and task domains. Below, we discuss limitations, open design choices, and future directions.

\textbf{Syntax Error Handling.} When tokenizing the code, we use AST-based parsing to accurately extract code components, e.g., comments. However, AST-based parsing cannot handle code with syntax errors. Our pipeline handles these cases gracefully: samples that fail to parse contribute zero tokens to the frequency analysis. In practice, the proportion of unparseable samples is small, as all 10 LLMs are highly capable. This strategy therefore does not significantly impact our results.

\setlength{\belowcaptionskip}{-12pt}
\begin{figure}[tb]
  \centering 
  \includegraphics[width=\columnwidth]{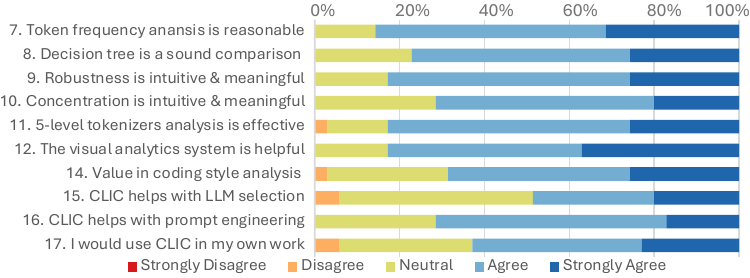}
  \vspace{-0.22in}
  \caption{Post-session survey results for 10 key questions.}
  \label{fig:survey}
\end{figure}
\setlength{\belowcaptionskip}{0pt}

\textbf{Online Retraining of Decision Trees.} In the \treeview{}, users can interactively remove a feature and trigger an immediate retraining of the decision tree with $n{-}1$ features. At the current scale, retraining completes in seconds and introduces negligible latency. 
When computing the robustness and concentration metrics, we have also precomputed all trees offline; the cache mechanism (Sec.~\ref{sec:ablation_view}) restores any step instantly, eliminating wait time during interactive explorations.

\textbf{Hyperparameters.} Three hyperparameters govern our analysis pipeline: the feature vector length (the 500 most frequent tokens), the number of features used in the entropy calculation (top-15), and the step fraction used to compute concentration ($1/2$ of $S$ in Eq.~\ref{eq:concentration}). These values were determined empirically through iterative exploration but are not unique. For instance, concentration could be accumulated over the first $2/3$ of steps instead of $1/2$. Such choices do not affect the overall conclusions; as long as all LLM pairs are evaluated under the same hyperparameter settings, the resulting metrics remain fully comparable.

\textbf{Beyond Token Frequencies and Python Tasks.} \sysname{} characterizes LLM coding behavior through token frequencies for four reasons justified in Sec.~\ref{sec:introduction}. We acknowledge, however, that token frequencies do not directly model \textit{structured properties of programs} such as code length, syntactic complexity, or control-flow shape. Extending the analytical pipeline to incorporate per-sample code-length normalization (TF-IDF~\cite{sparck1972statistical}), AST-based structural features, or control-flow graphs is therefore a natural future direction. A second, complementary axis of extension concerns the \emph{task domain}: our studies are grounded in Python code generation for Kaggle competitions. While the analytical pipeline of \sysname{} is task-agnostic, an empirical characterization of how behavioral fingerprints differ across other software-development scenarios---API integration or multi-file maintenance---would meaningfully broaden our work. Both axes can be pursued without modifying the rest of the framework: \sysname{}'s modular pipeline provides a clean foundation for such extensions.

Additionally, the behavioral fingerprints produced by \sysname{} open several compelling directions beyond pairwise comparison.
\emph{First}, the fine-grained style signals provide a forensic lens on model lineage. A model that has been distilled from a teacher model should closely replicate the teacher's coding patterns and remain nearly indistinguishable inside \sysname{}. This makes \sysname{} a practical tool for surfacing undisclosed distillation relationships.
\emph{Second}, practitioners currently choose LLMs based largely on availability, reputation, or personal familiarity---there is little actionable guidance on \emph{how} LLMs actually differ. \sysname{} constructs a \emph{behavioral persona} for each LLM, giving users a principled basis for model selection.
\emph{Third}, the discriminative tokens surfaced by \sysname{} identify precisely where two LLMs diverge, providing direct, evidence-based guidance for prompt engineering. Users can leverage these signals to steer code generation toward a desired coding style.

%% file: tex/10conclusion.tex
\section{Conclusion}
\label{sec:conclusion}
In this paper, we present \sysname{}, a visual analytics approach for comparing LLM coding behaviors through token-frequency analysis. \sysname{} trains an interpretable decision tree on token-frequency vectors across five tokenization levels and characterizes the separability between two LLMs into four regimes based on two orthogonal metrics: robustness and concentration. Case studies comparing code across LLMs, tokenization levels, LLM temperatures, and LLM prompts demonstrate the practical value of the approach. We hope our work encourages the community to look beyond performance-based LLM evaluation and adopt behavior-based characterization as a complementary lens.